\documentclass[pdflatex,sn-mathphys-num]{sn-jnl}

\usepackage{graphicx}%
\usepackage{multirow}%
\usepackage{amsmath,amssymb,amsfonts}%
\usepackage{amsthm}%
\usepackage{mathrsfs}%
\usepackage[title]{appendix}%
\usepackage[table]{xcolor}%
\usepackage{textcomp}%
\usepackage{manyfoot}%
\usepackage{booktabs}%
\usepackage{algorithm}%
\usepackage{algorithmicx}%
\usepackage{algpseudocode}%
\usepackage{listings}%

\usepackage{pifont}
\usepackage{bm}
\usepackage{float}

\usepackage[final]{changes}

\usepackage{tabu}

\usepackage{fancyhdr}

\fancypagestyle{firstpage}{
    \fancyhf{}
    \fancyfoot[C]{\small Accepted by \textit{Cognitive Neurodynamics}, 2026. DOI: 10.1007/s11571-026-10574-5}

}

\theoremstyle{thmstyleone}%
\theoremstyle{thmstyletwo}%

\theoremstyle{thmstylethree}%

\begin{document}
\thispagestyle{firstpage}

\makeatletter
\newcommand*{\@rowstyle}{}
\newcommand*{\rowstyle}[1]{\gdef\@rowstyle{#1}\@rowstyle\ignorespaces}
\newcolumntype{=}{>{\gdef\@rowstyle{}}} 
\newcolumntype{+}{>{\@rowstyle}}        
\makeatother

\title[Article Title]{Sleep Stage Classification across Heterogeneous EEG/EOG Channels with Multi-Channel Differential Transformer}


\author[1]{\fnm{Benjamin Wei Hao} \sur{Chin}}\equalcont{These authors contributed equally to this work.}

\author[2]{\fnm{Yuin Torng} \sur{Yew}}\equalcont{These authors contributed equally to this work.}

\author[1,3]{\fnm{Jia Hong} \sur{Puah}}\equalcont{These authors contributed equally to this work.}

\author[1]{\fnm{Haocheng} \sur{Wu}}

\author[1]{\fnm{Lanxin} \sur{Liang}}

\author*[2]{\fnm{Chow Khuen} \sur{Chan}}\email{ckchan@um.edu.my}

\author[2]{\fnm{Norita Mohd} \sur{Zain}}

\author[4]{\fnm{Siti Balqis} \sur{Samdin}}
\author*[1]{\fnm{Sim Kuan}\sur{Goh}}\email{simkuan.goh@xmu.edu.my}

\affil*[1]{\orgdiv{School of Artificial Intelligence and Robotics}, \orgname{Xiamen University Malaysia}, \orgaddress{ 43900 Sepang, \state{Selangor}, \country{Malaysia}}}

\affil[2]{\orgdiv{Department of Biomedical Engineering}, \orgname{Universiti Malaya}, \orgaddress{Kuala Lumpur 50603, \country{Malaysia}}}

\affil[3]{\orgdiv{School of Energy and Chemical Engineering}, \orgname{Xiamen University Malaysia}, \orgaddress{Sepang 43900, \state{Selangor}, \country{Malaysia}}}

\affil[4]{\orgdiv{Department of Biomedical Engineering and Health Sciences}, \orgname{Universiti Teknologi Malaysia}, \orgaddress{Johor Bahru 81310, \state{Johor}, \country{Malaysia}}}


\abstract{
Classification of sleep stages is essential for assessing sleep quality and diagnosing sleep disorders. However, manual inspection of EEG characteristics for each stage is time-consuming and prone to human error. Although machine learning and deep learning methods have been actively developed, they continue to face challenges arising from the non-stationarity and variability of electroencephalography (EEG) and electrooculography (EOG) signals across diverse clinical configurations, often resulting in poor generalization. In this work, we propose SleepDIFFormer, a multi-channel differential transformer framework for heterogeneous EEG–EOG representation learning. SleepDIFFormer is trained across multiple sleep staging datasets, each treated as a source domain, with the goal of generalizing to unseen target domains. Specifically, it employs a Multi-channel Differential Transformer Architecture (MDTA) designed to process raw EEG and EOG signals while incorporating cross-domain alignment. Our approach mitigates spatial and temporal attention noise and learns a domain-invariant EEG–EOG representation through feature distribution alignment across datasets, thereby enhancing generalization to new domains. Empirically, we evaluated SleepDIFFormer on five diverse sleep staging datasets under domain generalization settings and benchmarked it against existing approaches, achieving superior performance. Furthermore, comprehensive ablation analyses and interpretation of the differential attention weights demonstrated their relevance to learning discriminative sleep EEG representations. These findings support the development of more generalizable approaches to automated sleep-stage classification. Our source code and checkpoint are made publicly available at \url{https://github.com/Ben1001409/SleepDIFFormer}
}




\maketitle

\section{Introduction}
Sleep is a critical biological process that ensures immune regulation, supports memory consolidation, and maintains metabolic stability. Disruption of sleep patterns has been shown to degrade cognitive performance, particularly memory retention and attentional control \cite{song2023total}, while increasing susceptibility to emotional instability \cite{goldstein2014role,yew2024eeg}. Recent epidemiological studies report that the prevalence of diagnosed sleep disorders nearly doubled between 2011 and 2020 \cite{ahn2024elevated}, underscoring the growing need for reliable and scalable computational approaches for sleep monitoring and disorder detection. Sleep stage classification (also referred to as sleep scoring) constitutes the foundation for subsequent diagnosis and treatment of sleep disorders \cite{aboalayon2016sleep}. In clinical practice, polysomnography (PSG) remains the gold standard for sleep assessment, capturing multiple physiological modalities such as electroencephalogram (EEG), electrooculogram (EOG), electrocardiogram (ECG), and electromyogram (EMG) \cite{keenan2005overview}. According to Rechtschaffen and Kales (R\&K) rules \cite{Rechtschaffen_1968}, PSG signals are typically classified into six sleep stages: wakefulness (Wake), non-rapid eye movement (NREM) stages (S1, S2, S3, S4), and rapid eye movement (REM). 

Among PSG measurements, EEG and EOG signals provide spatio-spectral-temporal characteristics that allow experts to identify subtle changes in sleep stages \cite{shustak2019home}. For instance, sleep spindles in the 12–16 Hz range are present in stage N2, whereas delta waves around 0.5–4 Hz are present in stage N3. EOG detects slow rolling eye movements typically observed in stage N1, as well as rapid bursts characteristic of REM stage. Given the typical duration of overnight sleep recordings, manual inspection is time-consuming and prone to human errors. While machine learning and deep learning methods have been actively developed, they continue to face challenges because EEG and EOG signals are inherently noisy due to their non-stationarity and high variability across domains (i.e., datasets). Additional challenges include artifacts and diverse electrode placement \cite{jiang2019removal,goh2017automatic}. Hence, sleep stage classification demands methods that can learn more robust joint EEG and EOG representations.

In clinical settings, sleep staging datasets are often collected under varying conditions, including differences in acquisition equipment, electrode placement, and referencing, which result in significant distributional shifts across domains (i.e., datasets) and pose major challenges for generalization. \cite{wang2024generalizable} investigated domain generalization using a convolutional architecture as the backbone with a multi-level domain alignment loss function, and has shown promising results. Meanwhile, transformer models have gained attention from researchers in language~\cite{achiam2023gpt}, vision~\cite{dosovitskiy2020image}, and time-series forecasting \cite{nie2022time} for the capability of learning dependencies underlying time series segments after patching. Recent studies have demonstrated that transformer-based models are more effective at capturing multivariate time-series representations \cite{wang2024timexer}, by introducing an architecture for representing endogenous and exogenous variables.
Hence, transformers could further improve robustness and generalization for EEG signals that are inherently more complex. 
Moreover, transformers have also been building blocks for foundation models (FMs), which learn general representations from multiple time-series datasets for zero-shot forecasting or task-specific fine-tuning~\cite{liu2025sundial}. In the context of EEG, FMs have been trained on numerous EEG datasets to obtain generalizable representations~\cite{jiang2024large,wang2025cbramod}. However, empirical studies have shown that performance improvements are marginal with increasing model scale, while the computational cost remains prohibitively high. Moreover, zero-shot generalization performance has been found to be challenging, requiring further investigation and continued research~\cite{jiang2025neurolm,elmoaqet2022deep,van2024deep,aziz2025bcinetv1}. Here, we attempt to address these challenges to enable stronger generalization for clinical utility.

In this work, we propose a Sleep Stage Classification method by developing Multi-channel Differential Transformer (SleepDIFFormer) for heterogeneous EEG and EOG representation learning. Specifically, SleepDIFFormer comprises Multi-channel Differential Transformer Architecture (MDTA), combined with cross-domain feature alignment. The MDTA mitigated signal and attention noise, while the cross-domain feature alignment learned domain-invariant features through feature distribution alignment across datasets, enabling generalization to unseen target datasets. Empirically, we evaluated SleepDIFFormer on five different sleep staging datasets and conducted comprehensive ablation analyses to assess the contribution of each component.

The main contributions of this work are as follows:
\begin{enumerate}
\item We propose SleepDIFFormer, a framework that captures heterogeneous EEG–EOG representations across diverse clinical configurations for sleep stage classification.
\item We developed the MDTA to learn hierarchical spatio-temporal dependencies of EEG and EOG, while suppressing inherent signal \& attention noise.
\item We incorporated cross-domain feature alignment across datasets during training to enhance generalization.
\item We evaluated our method on five public datasets, compared strong baselines, and conducted comprehensive ablation studies.
\end{enumerate}

The remainder of this paper is organized as follows. Section 2 provides a review of related work. Section 3 describes the proposed method, SleepDIFFormer. Section 4 presents the experimental setup, empirical results, and comparative analysis against existing methods, followed by an ablation study. Section 5 discusses the results and interprets the representations learned by our method. Finally, Section 6 concludes the paper with a summary and directions for future research.

\section{Related Works}

Existing literature was organized based on the methodological evolution from traditional machine learning to advanced deep learning, followed by the potential of domain generalization techniques and transformer models. Early machine learning approaches for sleep staging relied on handcrafted EEG and EOG feature engineering (e.g., statistical descriptors~\cite{jiang2021robust}, and graph-based measures~\cite{Jain_Ganesan_2021}). However, their performance often plateaued and did not scale effectively with larger datasets. Deep learning introduced a paradigm shift from manual feature engineering to end-to-end neural architecture learning, in which network modules are designed to automatically capture highly relevant aspects of the data. Widely used architectures include convolutional neural networks (CNNs)~\cite{goh2018spatio,perslev2019u,jia2021salientsleepnet}, recurrent neural networks (RNNs)~\cite{Supratak_Dong_Wu_Guo_2017c,9176741,9746353} and graph neural networks (GNNs)~\cite{jia2020graphsleepnet,10095397}, each capturing distinct characteristics of the EEG and EOG data, including local spatial patterns, temporal dependencies, and relational structures. Furthermore, a number of hybrid models have been introduced to integrate the complementary architectural strengths for more comprehensive representational capacity~\cite{9746353,qu2020residual,aziz2025bcinetv1}. However, these works were primarily trained and validated on the same dataset, without examining generalization to other datasets collected under different clinical settings.

For clinical applications, ensuring the transferability of models to unseen datasets is of critical importance, as EEG and EOG signals are typically recorded under diverse clinical configurations. Considerable efforts have focused on domain adaptation techniques, which aim to reduce distribution gaps between training and validation domains in the absence of labeled data \cite{eldele2022adastattentivecrossdomaineegbased, zhou2025personalized}, with partially labeled data \cite{semisupervisedDA, xia2024eegnet}. To handle inter-subject variability, subject-adaptation was adopted in~\cite{wang2024subject,9530406}. In contrast, domain generalization seeks to develop models that can generalize to previously unseen target domains. SleepDG \cite{wang2024generalizable} represents a notable framework addressing domain shifts across datasets by employing a convolutional backbone trained with supervised learning and multi-level domain alignment. This work laid the groundwork for models to generalize to new data distributions without further training or fine-tuning. Our work builds upon this line of research by introducing a dedicated architectural design that explicitly accounts for the inherent dependencies and noise present in EEG/EOG signals.

Transformer models, equipped with self-attention and cross-attention mechanisms, enable neural networks to capture dependencies within a single input and across multiple inputs. In~\cite{eldele2021attention} and \cite{he2025multilevel},  multi-resolution CNN features were fused using multi-head attention for integrating local motifs and global context, while~\cite{phan2022sleeptransformer} demonstrated its competitiveness against CNN–RNN hybrids. Recent studies in EEG have adapted transformer to extract multi-channel time–frequency representations for automatic sleep staging~\cite{daimultichannelsleepnet} and emotion recognition~\cite{lin2025mbrstcformer, wang2025mtfsfn}. In~\cite{Guo_FlexSleepTransformer}, a transformer-based approach was proposed to handle multiple channels of raw EEG signals, where channel-specific encoders were employed and their outputs were subsequently aggregated and fed into a sequence encoder to form a joint feature representation. Transformer-based architectures have recently been extended to FMs~\cite{yang2023biot,jiang2024large}, which employ self-supervised learning across multiple datasets and are subsequently fine-tuned for specific downstream tasks. While FMs generally outperform non-FMs trained on a single dataset, the performance improvements tend to plateau with increasing model scale, and the associated computational cost becomes prohibitive~\cite{wang2024eegpt,jiang2025neurolm,wang2025cbramod}. Furthermore, although~\cite{puah2025eegdmeegrepresentationlearning} demonstrated that non-FM diffusion models can surpass existing FMs, these models still require fine-tuning on individual datasets and fail to explicitly address domain generalization to unseen datasets. Moreover, two other challenges remain when applying transformer models to EEG: (i) coping with inherent noise in both the signals and the attention mechanism, and (ii) adapting the architecture to effectively handle multivariate inputs (e.g., EEG and EOG) with domain generalization. In this work, we address these challenges by leveraging insights from time-series transformer models~\cite{wang2024timexer} and differential transformer architectures~\cite{ye2024differential}, originally developed for language modeling with attention noise reduction, and adapting them to model the inherent dependencies in EEG/EOG signals while suppressing noise.

\section{Method} 
This section provides a detailed description of the SleepDIFFormer architecture, outlining its core components, design principles, and functional building blocks.
\subsection{Main Framework} 
\begin{figure}
 \centering
 \includegraphics[width=\textwidth, trim=20 24 16 25, clip] {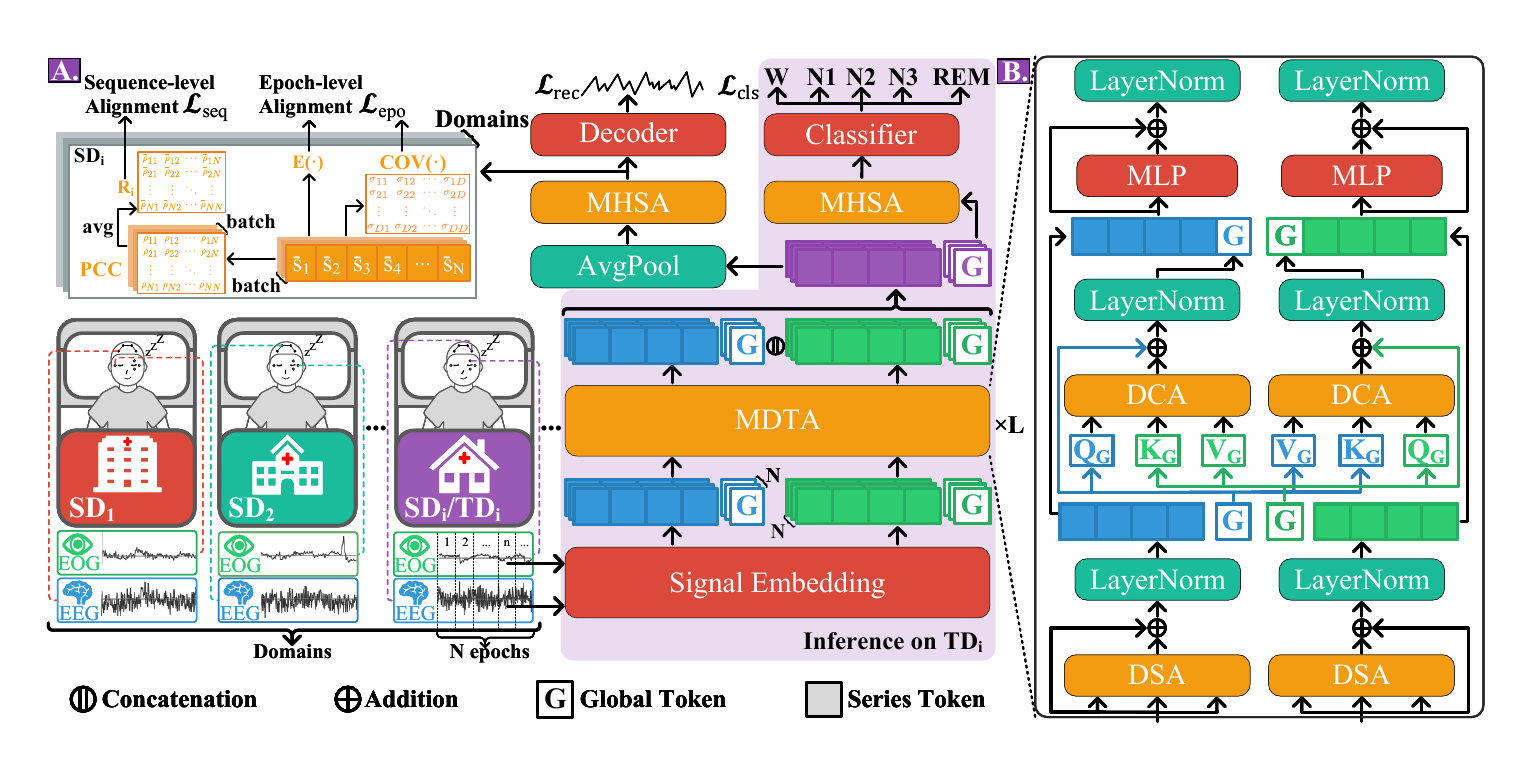}
    \caption{\textbf{Panel A} illustrates the overall architecture of SleepDIFFormer, which is trained on heterogeneous EEG and EOG data from multiple source domains $SD_{\{\cdot\}}$ (collected across different hospitals and recording conditions). To evaluate its cross-domain generalization capability, the trained model is directly tested on a completely unseen target domain $TD_i \notin SD_{\{\cdot\}}$ in a strictly zero-shot manner, without any parameter updates or test-time adaptation. \textbf{Panel B} provides a zoomed-in view of the Multi-channel Differential Transformer Architecture (MDTA), which comprises our proposed differential self-attention (DSA) and differential cross-attention (DCA).}
    \label{fig:main}
\label{general}
\end{figure}

We illustrate our main framework, SleepDIFFormer, in Fig.~\ref{fig:main}, which comprises signal embedding, MDTA, and a module for capturing inter-epoch dependencies. Our method was developed to learn domain-invariant heterogeneous EEG–EOG representations for sleep stage classification. SleepDIFFormer was trained using EEG and EOG data from multiple source domains. Here, $SD_i$ denotes the $i$-th source domain where each sample consists of $N$ epochs, and the model was optimized using data from multiple source domains $SD_{\{\cdot\}}$ with the objective of generalizing effectively to an unseen target domain, denoted as $TD_i \notin SD_{\{\cdot\}}$.

\subsection{Signal Embedding}  
To capture temporal variations in raw EEG and EOG, we designed a hierarchical convolutional module for signal embedding (SE), which converts raw signals into tokens, as shown in Fig.~\ref{fig:main}.
\begin{align}
\{s_1, s_2, \ldots, s_N\} &= \phi(x)
\end{align}
Here, \( x = \{x_1, x_2, \dots, x_T\} \in \mathbb{R}^{T \times 1} \) denotes the raw input with $T$ samples, where \( \phi(\cdot) \) denotes signal embedding module consists of three 1D convolutional blocks that extract hierarchical temporal features from the input signal with increasing the channel dimensionality from 1 to 64, then 128, and finally 128. The first convolution uses a larger kernel size of 49 to capture temporal patterns, followed by two smaller convolutions with kernel sizes of 7 and 5 for finer feature extraction. Each convolution is followed by Batch Normalization, a GELU activation function, and MaxPool1D for temporal downsampling, with pooling kernel/stride sizes of 5, 5, and 4, respectively. 
After passing through the \( \phi(\cdot) \), a set of \( N \) series tokens \( \{s_1, \dots, s_n, \dots, s_N\}\) were obtained, with $s_n \in \mathbb{R}^{d}$.

The resulting tokens were further enriched by adding learnable positional embeddings (PE), applied separately to modality $m$, which is EEG or EOG.
\begin{align}
\text{S}_m &= \text{PE}(\{s_1, s_2, \dots, s_N\})
\end{align}

\subsection{MDTA} 
In multi-head attention (MHA), the query ($Q$), key ($K$), and value ($V$) representations used to compute attention are obtained by projecting their respective input sources $X_Q, X_K, X_V$ through learned weight matrices:
\begin{align}
Q &= X_Q W_Q, \quad
K = X_K W_K, \quad
V = X_V W_V,
\end{align}
where $W_Q, W_K, W_V$ are trainable parameters. The attention operation is then defined as
\begin{align}
\operatorname{MHA}(Q,K,V) &= 
\sigma\left(\tfrac{QK^\top}{\sqrt{d_k}}\right)V.
\end{align}
where $\sigma$ is the Softmax function. A typical special case is multi-head self-attention (MHSA), where all three inputs are identical ($X_Q = X_K = X_V$). Another is multi-head cross-attention, where the query originates from one source ($X_Q$) and the key–value pairs from another ($X_K = X_V$).

Differential Transformers (DT) \cite{ye2024differential} built upon multi-head attention (MHA) and enhanced robustness to attention noise through the Differential Attention (DA) mechanism, originally proposed for language modeling tasks, denoted as $\mathrm{DiffAttn}(Q,K,V)$.

\begin{align}
\mathrm{DiffAttn}(\cdot)
= \big(\sigma(\frac{Q_1K_1^T}{\sqrt{d'}}) - \lambda\,\sigma(\frac{Q_2K_2^T}{\sqrt{d'}})\big)V
\end{align}

Specifically, DA constructs two sets of MHSA modules by decomposing the query, key, and value representations. Specifically, the query $Q$ is split into two parts $Q_1$ and $Q_2$, the key $K$ into $K_1$ and $K_2$, while the value $V$ remains unchanged. Subsequently, it computes the weighted difference with $\lambda$ between the attention scores from two parts. $d'$ is the dimension for tokens. This mechanism amplifies important signal components and cancels noise, thereby improving both the sparsity and selectivity of the attention distribution. The weighting scalar $\lambda$ serves as a weighting factor that modulates the subtraction of the secondary attention map, directly controlling the noise-suppression properties of the mechanism. Following the design \cite{ye2024differential}, $\lambda$ is implemented as a learnable parameter. It is initialized to $0.8$ for stable gradient flow during the early stages of training. Throughout the learning process, $\lambda$ is updated via backpropagation, allowing the model to dynamically refine its scaling factor and adaptively suppress noise based on the specific statistics and heterogeneity of the input EEG/EOG channels. Our MDTA, illustrated in Fig.~\ref{general}, comprises:

\subsubsection{Differential Self-attention (DSA) for Intra-Channel } 
We extended DA to represent EEG and EOG signals. For each modality, we introduced a learnable global token \( G \) that functions as a semantic hub. During attention computation, this token interacted with tokens within modalities to integrate features across the full sequence and can subsequently be used to fuse semantic representations between modalities.

The DSA mechanism is defined as follows:
\begin{align}
S_m',G_m' &=\text{DSA}([S_m,G_m])\\
S_m'',G_m''& = \text{LayerNorm}([S_m,G_m] +[S_m',G_m'])
\end{align}
The concatenated series tokens $S_m$ and global tokens $G_m$ serves as the shared input to construct $Q, K, V$ for differential self-attention module. The output was then integrated with the original input via a residual connection, followed by layer normalization.

\subsubsection{Differential Cross-attention (DCA) for Inter-Channel } 

To capture multi-channel representations and interactions, DA is incorporated into cross-attention mechanism between the global tokens of EEG and EOG, facilitating feature fusion while suppressing noise.
\begin{align}
G_{\text{eeg}}''' &= \text{LayerNorm}\left(G_{\text{eeg}}'' + \text{DCA}(G_{\text{eeg}}'',\ G_{\text{eog}}'')\right) \\
G_{\text{eog}}''' &= \text{LayerNorm}\left(G_{\text{eog}}'' + \text{DCA}(G_{\text{eog}}'',\ G_{\text{eeg}}'')\right)
\label{eq:Cross_Diffattn}
\end{align}
The global tokens $G_{\text{eeg}}'''$ and $G_{\text{eog}}'''$ are obtained via differential cross-attention $\text{DCA}(G''_m,G''_{m'})$, where the same modality $m$ serves as $Q$ and the other modality $m'$ provides $K$ and $V$, with the outputs passed through MLP projection and residual addition, followed by layer normalization.
\begin{align}
\tilde{S}_m,\ \tilde{G}_m &= \text{LayerNorm}(\text{MLP}([S_m'',G_m'''])+{[S_m'',G_m''']})
\label{eq:conv1d_projection}
\end{align}
where $\tilde{S}_m$ and $\tilde{G}_m$ represent the series tokens and learned global token for EEG or EOG modality. 
\subsection{MHSA for Inter-Epoch Dependencies} 
After 
$L$ layers of MDTA, 
MHSA was employed to model sequence-level context and enhance the model’s ability to capture the evolution of sleep stages over long periods. In contrast to local modeling within a single epoch, sequence-level encoding emphasizes phase continuity and temporal trends across successive epochs. Both the standard MHSA and DA can be applied here. However, as shown in the ablation study in Section~\ref{sec:mainresult}, DA negatively impacted performance when used for inter-epoch dependencies, likely due to its tendency to amplify inter-epoch differences while suppressing intra-sequence common structures. Therefore, we utilized MHSA with concatenation (Concat) and global average pooling (AvgPool):
\begin{align}
\bar{G} &= \text{MHSA}(\text{Concat}(\tilde{G}_{\text{eeg}},\ \tilde{G}_{\text{eog}})) \label{eq:G} \\
\bar{S} &= \text{MHSA}(\text{AvgPool}(\text{Concat}(\tilde{S}_{\text{eeg}},\ \tilde{S}_{\text{eog}})))
\label{eq:sequence_encoder_inter_epoch}
\end{align}
where $\tilde{G}_{\text{eeg}}$ and $\tilde{G}_{\text{eog}}$ denote the global tokens of each EEG and EOG epoch, respectively. $\tilde{S}_{\text{eeg}}$ and $\tilde{S}_{\text{eog}}$ represent the corresponding series tokens, 
$\bar{S}$ is the fused feature map and $\bar{G}$ is the fused global token with dimensionality 
$\mathbb{R}^{D=2d}$.

\subsection{Loss Function}

The $\bar{G}$ from Equation~\ref{eq:G} was passed to a linear classifier for sleep stage classification with cross-entropy loss:
\begin{equation}
\mathcal{L}_{\text{cls}} = - \sum_{n=1}^N \sum_{c=1}^C y_{n,c}^{SD_i} \log\left( \mathrm{Pred}_{n,c}^{SD_i} \right)
\label{eq:classification_loss}
\end{equation}
where $C$ is the number of sleep stages and $N$ is the total number of epochs in the sequence from domain $SD_i$, $\mathrm{Pred}_{n,c}^{SD_i}$ denotes the predicted probability and $y_{n,c}^{SD_i}$ is the annotation for the $n$-th sample and $c$-th class from domain $SD_i$.

From Equation~\ref{eq:sequence_encoder_inter_epoch}, the $\bar{S}$ is passed to a decoder $\varphi(\cdot)$ (i.e., 5 layers CNN) for signal reconstruction of inputs. The reconstruction loss is defined as:
\begin{equation}
\mathcal{L}_{\text{rec}} = \frac{1}{T} \sum_{t=1}^{T} \left\| x_t - \hat{x}_t \right\|_2^2
\label{eq:reconstruction_loss}
\end{equation}
where \( x_t \) denotes the raw signal and \( \hat{x}_t \) denotes the reconstructed signal from $\varphi(\cdot)$ for each sleep epoch. \( \| \cdot \|_2^2 \) is the squared L2 norm. 

\begin{table}
\caption{\added{Summary of Datasets Studied, Characteristics, and Sleep Stage Distribution}}
\centering

\resizebox{\textwidth}{!}{
\begin{minipage}{\textwidth}
    \begin{tabular}{lccccccccc}
    \toprule
    \textbf{Domain $SD_i/TD_i$} & \textbf{\# of Subs} & \textbf{\# of Samples} & \textbf{Sampling Freq.} & \textbf{W(\%)} & \textbf{N1(\%)} & \textbf{N2(\%)} & \textbf{N3(\%)} & \textbf{REM(\%)} \\
    \midrule
    1. SleepEDFx & 197 & 236120 & 100 Hz  & 29.02 & 10.64 & 37.63 & 8.24 & 14.47 \\
    2. HMC & 154 & \added{135820} & 256 Hz & \added{17.00} & \added{11.27} & \added{36.61} & \added{19.59} &  \added{15.53} & \\
    3. ISRUC & 126 & 112060 & 200 Hz & 22.01 & 13.17 & 31.85 & 19.69 & 13.28 \\
    4. SHHS-1 & 5793 & 132720 & 125 Hz & 30.37  & 3.19 & 40.79 & 13.01 & 12.64 \\
    5. P2018 & 994 & 147340 & 200 Hz & 17.36  & 13.44  & 42.54 & 13.27 & 13.39 \\
    
    \bottomrule
    \end{tabular}
    
    \end{minipage}}
\label{table:styled_summary}
\end{table}

To improve cross-domain generalization, we incorporated a multi-level feature alignment (FA) loss that explicitly minimizes distributional difference between source domains~\cite{wang2024generalizable}. By aligning epoch-level statistical properties, mean \cite{mmd2012} and covariance \cite{sun2016correlationalignmentunsuperviseddomain}, together with sequence-level \cite{wang2024generalizable} statistical properties, the model was encouraged to learn domain-invariant representations while preserving temporal structures.
The total epoch-level alignment loss consists of two components:
\begin{align}
\mathcal{L}_{\text{exp}}&=\sum_{i \neq j} \left\| \mathbb{E}(\bar {S_i}) - \mathbb{E}(\bar {S_j}) \right\|_2^2\\
\mathcal{L}_{\text{cov}}&=\sum_{i \neq j} \left\| \mathrm{COV}(\bar {S_i}) - \mathrm{COV}(\bar {S_j}) \right\|_F^2\\
\mathcal{L}_{\text{epo}} &=\mathcal{L}_{\text{exp}}+\mathcal{L}_{\text{cov}}
\label{eq:epoch_level_alignment}
\end{align}
where \( \bar{S}_i \) and \( \bar{S}_j \) denote the feature sets from different source domains $SD_i$ and $SD_j$. \( \mathbb{E}(\cdot) \) and \( \mathrm{COV}(\cdot) \) represent the mean and covariance, while \( \| \cdot \|_2 \) and \( \| \cdot \|_F \) are the L2 and Frobenius norms operators. This loss enforced local distribution alignment across domains. To further promote inter-epoch alignment, we adopted:
\begin{equation}
\mathcal{L}_{\text{seq}} = \sum_{i \neq j} \left\| R_i - R_j \right\|_F^2
\label{eq:sequence_level_alignment}
\end{equation}
where \( R_i \) and \( R_j \) are correlation matrices computed using the Pearson Correlation Coefficient (PCC) from the sleep sequences in each domain $SD_i$ and  $SD_j$. This component captured and aligned the temporal dependency structure across domains at the sequence level.
The overall loss function:
\begin{equation}
\mathcal{L}_{\text{total}} = \mathcal{L}_{\text{cls}} + \lambda_{\text{rec}} \mathcal{L}_{\text{rec}} + \lambda_{\text{align}} \left( \mathcal{L}_{\text{epo}} + \mathcal{L}_{\text{seq}} \right)
\label{eq:total_loss}
\end{equation}
where the parameters \( \lambda_{\text{rec}} \) and \( \lambda_{\text{align}} \) control the relative importance of reconstruction and domain alignment.

\section{Experiment and Results} 
We provided detailed descriptions of the sleep EEG datasets, preprocessing steps, experimental settings, main results, comparative analysis, and ablation studies.

\subsection{Datasets \& Experimental Settings}

We utilized five publicly available sleep EEG datasets, summarized in Table~\ref{table:styled_summary}. 
We followed the American Academy of Sleep Medicine
(AASM) recommendations 
\cite{berry2014aasm} 
and merged the N3 and N4 stages in the SleepEDFx and SHHS datasets into a unified N3 stage. We follow the established protocol in SleepDG~\cite{wang2024generalizable}, selecting the first 150 recordings from the large-scale SHHS (5,793 recordings) and P2018 (994 recordings) datasets, as shown in Table~\ref{table:styled_summary}. This setting ensures a fair and standardized comparison while addressing the imbalance in dataset sizes. Following \cite{Supratak_Dong_Wu_Guo_2017c}, each recording in SleepEDFx was cropped to include only segments ranging from 30 minutes before the onset of the first sleep epoch to 30 minutes after the end of the final sleep stage, due to the long Wake stage. Our preprocessing includes bandpass filtering (0.3–35 Hz), resampling to 100 Hz, and Z-score normalization (channel-wise). For dataset SHHS-1, EEG signals were recorded using a referential montage. To ensure comparability with other datasets that used bipolar montage, we constructed a bipolar channel by subtracting two electrodes, which are the ROC and LOC electrodes, and used it as the EOG channel for the SHHS-1 dataset.

We followed the same experimental settings as described in \cite{wang2024generalizable}. In each experiment, four datasets were used for training and the remaining dataset for testing.  
We adopted the same EEG and EOG channel selection:  
\begin{enumerate}
    \item \textbf{SleepEDFx \cite{867928}}: EEG from Fpz--Cz and EOG from EOG--Horizontal.  
    \item \textbf{HMC \cite{HMC_Sleep_Staging_2021}}: EEG from F4--M1 and EOG from E1--M2.  
    \item \textbf{ISRUC \cite{Khalighi_Sousa_Santos_Nunes_2015d}} : EEG from F4--M1 and EOG from LOC--A2.  
    \item \textbf{SHHS-1 \cite{Zhang_Cui_Mueller_Tao_Kim_Rueschman_Mariani_Mobley_Redline_2018d}}: EEG from C4--M1 and EOG from ROC-LOC.  
    \item \textbf{P2018 \cite{Ghassemi_Moody_Lehman_Song_Li_Sun_Westover_Clifford_2018d}}: EEG from C3--M2 and EOG from E1--M2.  
\end{enumerate}

We set the length of the input sleep epoch sequence to $N = 20$, meaning that each input sample consists of 20 consecutive epochs, similar to \cite{wang2024generalizable}. The feature dimension, $d$, for each epoch was fixed at 128. Four layers ($L=4$) of MDTA were used during training, where each layer consisted of 4 heads. In the inter-epoch MHSA, only 1 layer was utilized, with 8 heads. As part of our training objective, we used a composite loss function with equal weights ($\lambda_{\text{rec}} = \lambda_{\text{align}} = 0.5$
) to ensure balanced optimization. The model performance was evaluated based on Accuracy and Macro-F1 score across 5 datasets.

Our code was implemented using PyTorch and the experiment was performed on a workstation with NVIDIA GeForce RTX 4090 GPU with 24GB VRAM. Training lasted for 50 epochs with a batch size of 16. The model was optimized using the Adam with a learning rate of $5 \times 10^{-4}$ and a dropout rate of 0.1 were applied.

\subsection{Main Results and Ablation Analyses}\label{sec:mainresult}

\begin{tableorg}[t]
\centering
\caption{\added[]{Comparative analysis of Generalization. Each method was trained using multiple source domains $SD_{\{\cdot\}}$ and evaluated on a target domain $TD_i \notin SD_{\{\cdot\}}$. The best results are highlighted in \textbf{bold}, while the second-best results are \underline{underlined}. $\dagger$ and $\star$ denote $p < 0.001$ for the paired $t$-test and the Wilcoxon signed-rank test, respectively.}}
\label{tab:cross_domain_sleep}

\setlength{\tabcolsep}{4pt}
\renewcommand{\arraystretch}{1.05}

\resizebox{\linewidth}{!}{%
\begin{tabular}{lcccccccccccc}
\toprule

\textbf{Source Domain $\bm{\mathrm{SD}_{\{\cdot\}}}$}
& \multicolumn{2}{c}{\textbf{$\bm{\mathrm{SD}_{\{2,3,4,5\}}}$}}
& \multicolumn{2}{c}{\textbf{$\bm{\mathrm{SD}_{\{1,3,4,5\}}}$}}
& \multicolumn{2}{c}{\textbf{$\bm{\mathrm{SD}_{\{1,2,4,5\}}}$}}
& \multicolumn{2}{c}{\textbf{$\bm{\mathrm{SD}_{\{1,2,3,5\}}}$}}
& \multicolumn{2}{c}{\textbf{$\bm{\mathrm{SD}_{\{1,2,3,4\}}}$}}
& \multicolumn{2}{c}{\textbf{Average}} \\
\midrule

\textbf{Target Domain $\bm{\mathrm{TD}_i}$}
& \multicolumn{2}{c}{\textbf{$\bm{\mathrm{TD}_1}$-SleepEDFx}}
& \multicolumn{2}{c}{\textbf{$\bm{\mathrm{TD}_2}$-HMC}}
& \multicolumn{2}{c}{\textbf{$\bm{\mathrm{TD}_3}$-ISRUC}}
& \multicolumn{2}{c}{\textbf{$\bm{\mathrm{TD}_4}$-SHHS-1}}
& \multicolumn{2}{c}{\textbf{$\bm{\mathrm{TD}_5}$-P2018}}
& \multicolumn{2}{c}{} \\
\midrule

\textbf{Metrics}
& ACC & MF1
& ACC & MF1
& ACC & MF1
& ACC & MF1
& ACC & MF1
& ACC & MF1 \\
\midrule

\multicolumn{13}{c}{\textbf{Reported Baseline Results}} \\
\midrule

DeepSleepNet \cite{Supratak_Dong_Wu_Guo_2017c}
& 72.28 & 65.72
& 64.04 & 62.84
& 69.71 & 67.80
& 64.73 & 52.79
& 66.62 & 60.51
& 67.48 & 61.93 \\

U-Time \cite{perslev2019u}
& 72.51 & 65.84
& 65.13 & 63.71
& 70.58 & 68.10
& 64.53 & 51.68
& 67.35 & 61.07
& 67.82 & 62.28 \\

AttnSleep \cite{eldele2021attention}
& 73.76 & 66.93
& 64.49 & 62.19
& 70.39 & 68.19
& 64.07 & 51.82
& 66.19 & 60.78
& 67.78 & 61.98 \\

ResnetMHA \cite{qu2020residual}
& 73.01 & 65.89
& 65.18 & 63.02
& 70.11 & 67.54
& 65.16 & 53.99
& 67.89 & 61.87
& 68.27 & 62.46 \\

TinySleepNet \cite{9176741}
& 73.34 & 66.10
& 65.87 & 64.01
& 69.18 & 66.99
& 65.76 & 54.56
& 68.29 & 61.36
& 68.49 & 62.60 \\

EnhancingCE \cite{9746353}
& 73.51 & 66.69
& 65.98 & 64.37
& 70.56 & 69.12
& 65.88 & 54.39
& 67.84 & 61.19
& 68.75 & 63.14 \\

SalientSleepNet \cite{jia2021salientsleepnet}
& 73.92 & 67.59
& 65.44 & 63.22
& 69.93 & 68.16
& 66.36 & 55.49
& 68.89 & 63.13
& 68.91 & 63.52 \\

SleepDG \cite{wang2024generalizable}
& \underline{77.44} & \underline{71.29}
& \underline{73.85} & \underline{71.16}
& \textbf{78.69} & \underline{74.44}
& {70.45} & {60.89}
& \underline{74.74} & \underline{70.43}
& \underline{75.03} & \underline{69.64} \\
\midrule
\multicolumn{13}{c}{\textbf{Reproduced Using Official Source Code}} \\
\midrule

{MSTGCN \cite{9530406}}
& {74.26} & {69.34}
& {67.86} & {65.20}
& {72.61} & {64.60}
& {\underline{73.58}} & {\underline{65.33}}
& {72.26} & {66.61}
& {72.11} & {66.22} \\

{LaBraM \cite{jiang2024large}}
& {66.34} & {55.31}
& {67.04} & {60.49}
& {58.59} & {53.14}
& {72.06} & {61.62}
& {70.22} & {62.59}
& {66.85} & {58.63} \\

{EEGPT \cite{wang2024eegpt}}
& {27.97} & {22.61}
& {59.42} & {50.31}
& {50.74} & {44.33}
& {65.68} & {54.11}
& {58.10} & {49.18}
& {52.38} & {44.11} \\
\midrule


\textbf{SleepDIFFormer (Ours)}
& \textbf{78.36} & \textbf{72.93}
& \textbf{75.00} & \textbf{72.71}
& \underline{78.08} & \textbf{75.47}
& \textbf{75.41} & \textbf{67.59}
& \textbf{74.95} & \textbf{72.09}
& \textbf{76.36} & \textbf{72.16} \\

\midrule
\multicolumn{13}{c}{\textbf{Paired $t$-test \& Wilcoxon signed-rank test between the Top 2 Methods}} \\
\midrule


SleepDG (ACC)
& \multicolumn{2}{l}{76.37 ± 0.72}
& \multicolumn{2}{l}{73.73 ± 0.53}
& \multicolumn{2}{l}{${77.80 \pm 0.37}^{\dagger\star}$}
& \multicolumn{2}{l}{74.90 ± 0.89}
& \multicolumn{2}{l}{73.42 ± 0.55}
& \multicolumn{2}{l}{75.24 ± 0.33} \\

SleepDG (MF1)
& \multicolumn{2}{l}{70.05 ± 0.77}
& \multicolumn{2}{l}{71.67 ± 0.48}
& \multicolumn{2}{l}{73.88 ± 0.55}
& \multicolumn{2}{l}{67.34 ± 0.72}
& \multicolumn{2}{l}{68.65 ± 0.60}
& \multicolumn{2}{l}{70.32 ± 0.32}
\\

\midrule

\textbf{SleepDIFFormer} (ACC)
& \multicolumn{2}{l}{${77.11 \pm 0.66}^{\dagger\star}$} 
& \multicolumn{2}{l}{${74.80 \pm 0.71}^{\dagger\star}$} 
& \multicolumn{2}{l}{77.40 ± 0.46} 
& \multicolumn{2}{l}{${75.15 \pm 0.87}$} 
& \multicolumn{2}{l}{${75.43 \pm 0.49}^{\dagger\star}$} 
& \multicolumn{2}{l}{${75.98 \pm 0.29}^{\dagger\star}$} 
\\

\textbf{SleepDIFFormer} (MF1)
& \multicolumn{2}{l}{${71.74 \pm 0.64 }^{\dagger\star}$} 
& \multicolumn{2}{l}{${72.55 \pm 0.63}^{\dagger\star}$}
& \multicolumn{2}{l}{${74.33 \pm 0.45}^{\dagger\star}$}
& \multicolumn{2}{l}{${67.16 \pm 0.76}$}
& \multicolumn{2}{l}{${72.41 \pm 0.42 }^{\dagger\star}$}
& \multicolumn{2}{l}{${71.64 \pm 0.26}^{\dagger\star}$} 
\\

\bottomrule

\end{tabular}%
}
\end{tableorg}

\begin{figure}
    \centering
    \includegraphics[width=\textwidth]{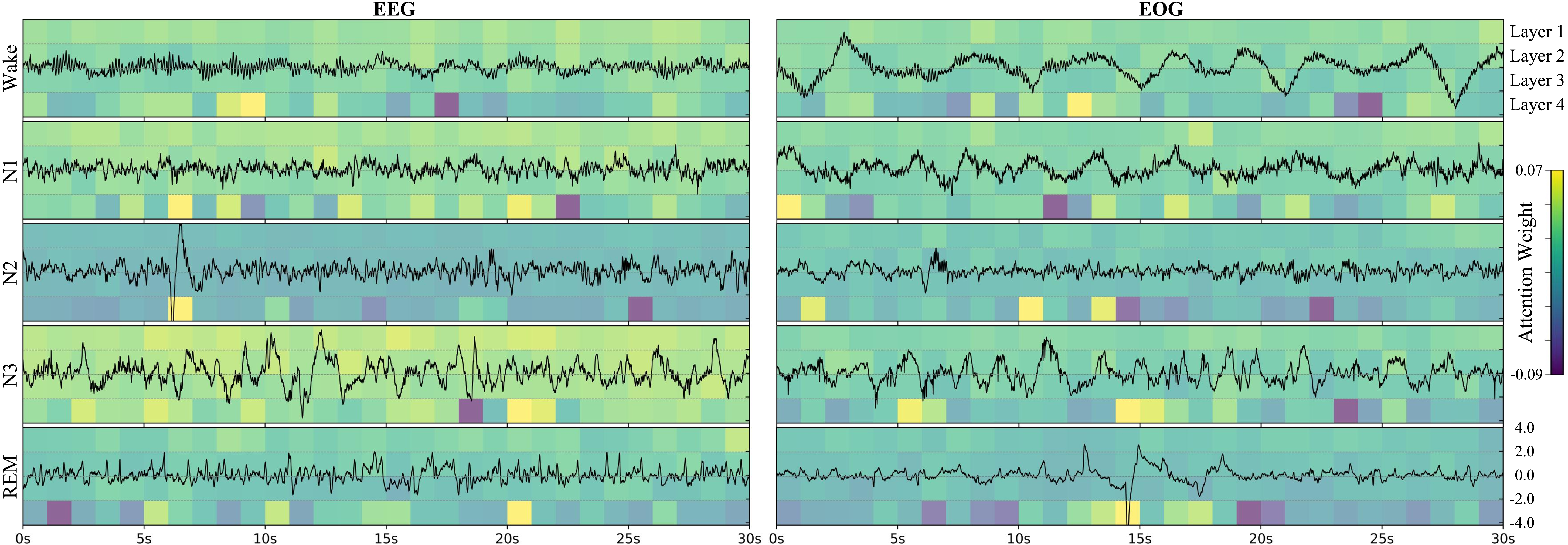}
    \caption{Layer-wise differential attention weights for EEG and EOG across different sleep stages. The attention maps are overlaid onto the corresponding waveforms to highlight stage-specific patterns. For visual clarity, the raw EEG and EOG signals were z-score normalized prior to plotting. Consistent color scale is used across all plots.}
    \label{fig:diffentialattention}
    \label{visual}
\end{figure}

\added[comment={R2 C2}]{We compared our method with the existing approaches listed in Table~\ref{tab:cross_domain_sleep}, using baseline results reported in the established literature and results reproduced using the official source code. We also conducted statistical tests between the two best-performing methods.} These results demonstrate SleepDIFFormer’s superior ability to capture underlying EEG representations that generalize to unseen datasets. The only exception was the ISRUC dataset, where SleepDIFFormer ranked second. This may be attributed to differences in the EOG channel configuration used in ISRUC compared with other datasets. Methods relying on feature engineering, such as using spectrograms in DeepSleepNet \cite{Supratak_Dong_Wu_Guo_2017c}, performed the worst, highlighting the advantage of models that learn representations directly from raw signals. SalientSleepNet \cite{jia2021salientsleepnet}, which adopted a $U^2$-CNN architecture, underperformed SleepDIFFormer, reinforcing the limitations of purely CNN-based models in addressing the non-stationarity of EEG and EOG signals. AttnSleep \cite{eldele2021attention}, ResnetMHA\cite{qu2020residual}, TinySleepNet \cite{9176741}, U-time \cite{perslev2019u} and EnhancingCE \cite{9746353} did not achieve notable improvements over the baseline DeepSleepNet \cite{Supratak_Dong_Wu_Guo_2017c}, despite extensive exploration of architectural variations. Additionally, MSTGCN \cite{9530406}, which integrated multi-view spatial-temporal graph networks with domain generalization, underperformed SleepDIFFormer. While MSTGCN successfully designs brain-region topologies for EEG, it lacks the architectural capacity to model cross-modal dynamics with EOG channels, which restricts its capability in capturing some comprehensive sleep stage features. \added[comment={R2 C7}]{We further evaluated two EEG FMs, LaBraM \cite{jiang2024large} and EEGPT \cite{wang2024eegpt}. EOG was included as an input channel during fine-tuning, consistent with the input configuration used for all evaluated methods. Despite their large-scale pretraining on diverse EEG datasets, including signals from frontal and prefrontal channels located near standard EOG electrode positions, LaBraM and EEGPT achieved lower cross-dataset performance than non-FM methods SleepDIFFormer, SleepDG, and MSTGCN. Their weaker cross-dataset performance may partly stem from their substantially larger parameter counts, which could increase susceptibility to overfitting to source-domain characteristics; however, this hypothesis requires further investigation.} Furthermore, \added[comment={R2 C1, R2 C2}]{we evaluated the statistical significance of the performance gains achieved by SleepDIFFormer over the second-best method, SleepDG. Specifically, we trained both models using 30 independent random seeds (0--29) and applied both the paired $t$-test and the Wilcoxon signed-rank test, with seed 0 also used for the main experiment. Overall, SleepDIFFormer achieved statistically significant ($p < 0.001$) improvements across most domains and evaluation metrics. The two methods exhibited comparable performance on SHHS-1, while SleepDG achieved higher ACC only on ISRUC.}

\begin{table}
\caption{\added{Ablation study of SleepDIFFormer. Ablating FA is equivalent to ablating both $\mathcal{L}_{\mathrm{epo}}$ and $\mathcal{L}_{\mathrm{seq}}$.}}

\centering
\begin{tabular}{cccccccccc|cc} 
\toprule 

\textbf{DA} & {\textbf{DSA}} & {\textbf{DCA}} & \textbf{SE} & \textbf{CA} & \textbf{FA} & \textbf{ID}  & $\mathcal{L}_{\text{rec}}$& $\mathcal{L}_{\text{epo}}$ & $\mathcal{L}_{\text{seq}}$ & \textbf{ACC} & \textbf{MF1} \\ 
\midrule 
\added{\ding{55}} & \added{\ding{55}} & \added{\ding{55}} & \added{\ding{51}} & \added{\ding{51}} & \added{\ding{51}} & \added{\ding{51}}  & \added{\ding{51}}  & \added{\ding{51}}  & \added{\ding{51}}  & \added{75.56} & \added{71.44} \\ 

{\ding{51}} & {\ding{55}} & {\ding{51}} & {\ding{51}} & {\ding{51}} & {\ding{51}} & {\ding{51}}  & \added{\ding{51}}  & \added{\ding{51}}  & \added{\ding{51}}  & {75.63 } & {71.34} \\ 

\added{\ding{51}} & \added{\ding{55}} & \added{\ding{51}} & \added{\ding{51}} & \added{\ding{51}} & \added{\ding{55}} & \added{\ding{51}}  & \added{\ding{51}}  & \added{\ding{55}} & \added{\ding{55}} & \added{75.71} & \added{71.54} \\

{\ding{51}} & {\ding{51}} & {\ding{55}} & {\ding{51}} & {\ding{51}} & {\ding{51}} & {\ding{51}}  & \added{\ding{51}}  & \added{\ding{51}}  & \added{\ding{51}}  & {76.18 } & {71.55 } \\ 

\added{\ding{51}} & \added{\ding{51}} & \added{\ding{55}} & \added{\ding{51}} & \added{\ding{51}} & \added{\ding{55}} & \added{\ding{51}}  & \added{\ding{51}}  & \added{\ding{55}} & \added{\ding{55}} & \added{76.19 } & \added{71.93 } \\

\ding{51} & {\ding{51}} & {\ding{51}} & \ding{55} & \ding{51} & \ding{51} & \ding{51}  & \added{\ding{51}}  & \added{\ding{51}}  & \added{\ding{51}}  & 73.40  & 68.65  \\ 
\ding{51} & {\ding{51}} & {\ding{51}} & \ding{51} & \ding{55} & \ding{51} & \ding{51}  & \added{\ding{51}}  & \added{\ding{51}}  & \added{\ding{51}}  & 75.67 & 71.22  \\ 
\ding{51} & {\ding{51}} & {\ding{51}} & \ding{51} & \ding{51} & \ding{51} & \ding{55}  & \added{\ding{51}}  & \added{\ding{51}}  & \added{\ding{51}}  & 76.05  & 71.38  \\
 \added{\ding{51}}  & \added{\ding{51}}  & \added{\ding{51}}  & \added{\ding{51}}  & \added{\ding{51}}  & \added{\ding{51}}  & \added{\ding{51}}  & \added{\ding{55}} & \added{\ding{51}}  & \added{\ding{51}}  & \added{75.97}&\added{71.79}\\
 \added{\ding{51}}  & \added{\ding{51}}  & \added{\ding{51}}  & \added{\ding{51}}  & \added{\ding{51}}  & \added{\ding{55}}& \added{\ding{51}}  & \added{\ding{51}}  & \added{\ding{55}} & \added{\ding{55}} & \added{76.13}&\added{71.78}\\
 \added{\ding{51}}  & \added{\ding{51}}  & \added{\ding{51}}  & \added{\ding{51}}  & \added{\ding{51}}  & \added{\ding{55}}& \added{\ding{51}}  & \added{\ding{55}} & \added{\ding{55}} & \added{\ding{55}} & \added{76.04}&\added{71.56}\\
 \added{\ding{51}}  & \added{\ding{51}}  & \added{\ding{51}}  & \added{\ding{51}}  & \added{\ding{51}}  & \added{\ding{51}}  & \added{\ding{51}}  & \added{\ding{51}}  & \added{\ding{55}} & \added{\ding{51}}  & \added{75.84}&\added{71.70}\\
 \added{\ding{51}}  & \added{\ding{51}}  & \added{\ding{51}}  & \added{\ding{51}}  & \added{\ding{51}}  & \added{\ding{51}}  & \added{\ding{51}}  & \added{\ding{51}}  & \added{\ding{51}}  & \added{\ding{55}} & \added{76.14}&\added{71.90}\\ 
\midrule \ding{51} & {\ding{51}} & {\ding{51}} & \ding{51} & \ding{51} & \ding{51} & \ding{51}  & \added{\ding{51}}  & \added{\ding{51}}  & \added{\ding{51}}  & \textbf{76.36} & \textbf{72.16} \\ 

\bottomrule \end{tabular}
\label{tab:ablation}
\end{table}

To examine the effectiveness of SleepDIFFormer in suppressing non-stationary noise and the ability of feature alignment in reducing domain shift, we conducted a leave-one-out ablation strategy to explicitly isolate the independent contribution of each component to the overall accuracy that is shown in Table~\ref{tab:ablation}, by removing DSA and DCA from MDTA (DA), individually or jointly, replacing signal embedding (SE) with patching, removing inter-channel cross-attention (CA), feature alignment (FA), using MHSA ($\checkmark$) vs.\ DA ($\times$) for inter-epoch dependency (ID)\added[comment={R2 C4}]{, and ablated the loss terms. Ablating FA is equivalent to ablating both $\mathcal{L}_{\mathrm{epo}}$ and $\mathcal{L}_{\mathrm{seq}}$.}

As shown in Table~\ref{tab:ablation}, removing both DSA and DCA reduced the average ACC and MF1, supporting their roles in suppressing non-stationary noise and enhancing salient features. Removing the signal embedding (SE) module caused the largest performance degradation, demonstrating the importance of effective and lightweight tokenization of raw signals. Excluding channel attention (CA) also reduced performance, suggesting that information exchange between EEG and EOG through global tokens benefits sleep-stage classification. Similarly, ablating ID decreased overall performance, supporting their respective contributions to learning domain-invariant representations and modeling temporal context across epochs. \added[comment={R2 C4}]{Ablating $\mathcal{L}_{\mathrm{rec}}$, $\mathcal{L}_{\mathrm{epo}}$, or $\mathcal{L}_{\mathrm{seq}}$ individually reduced both ACC and MF1, while jointly ablating all three resulted in the lowest MF1 among the loss-ablation configurations. The complete objective achieved the best overall performance, indicating that these loss terms provide complementary supervision.} \added[comment={R2 C5}]{The additional configurations further reveal interactions among DSA, DCA, and FA. Ablating FA when either DSA or DCA was absent produced a slight improvement; however, FA improved performance when both attention components were present. This suggests that FA is most effective when operating jointly with DSA and DCA. Overall, SleepDIFFormer achieved its best performance when all modules were included, indicating that the components provide complementary benefits.}

The hyperparameters were empirically selected through preliminary experiments on a small subset of the training data, based on validation performance observed over a few training epochs. To assess the sensitivity of hyperparameters in the loss function and MDTA architecture, we varied only one hyperparameter at a time and observed the impact on performance, as shown in Table~\ref{tab:hyperparameter_sensitivity}. The results show that the proposed framework remains stable across different configurations, with only moderate performance fluctuations, thereby validating the robustness of the proposed method.

\begin{table}
\centering
\caption{Sensitivity analysis of hyperparameters in the loss function and MDTA architecture. ACC and MF1 are averaged across five target domains under the leave-one-domain-out protocol. During each analysis, only the specified setting was varied, while all other settings were kept at their default values. The default hyperparameters are underlined, and the best results are highlighted in bold.
}
\label{tab:hyperparameter_sensitivity}

\setlength{\tabcolsep}{8pt}
\renewcommand{\arraystretch}{1.10}

\begin{tabular}{ccccc}
\toprule
\textbf{$\lambda_{\mathrm{cls}}$} &
\textbf{$\lambda_{\mathrm{rec}}$} &
\textbf{$\lambda_{\mathrm{align}}$} &
\textbf{Mean ACC} &
\textbf{Mean MF1} \\
\hline

\underline{1.00} &\underline{ 0.50 }& \underline{0.50}
& \textbf{76.36} & \textbf{72.16} \\

1.00 & 0.25 & 0.75
& 75.76 & 71.55 \\

1.00 & 0.75 & 0.25
& 75.57 & 71.46 \\

0.50 & 0.50 & 0.50
& 76.16 & 71.83 \\

1.00 & 0.25 & 0.25
& 75.73 & 71.50 \\

\midrule

\multicolumn{3}{c}{
\textit{Number of MDTA layers
}
} &\textbf{Mean ACC} &
\textbf{Mean MF1} \\
\hline

\multicolumn{3}{c}{\underline{4 layers}} & \textbf{76.36} & \textbf{72.16} \\
\multicolumn{3}{c}{2 layers} & 76.12 &	71.90 \\
\multicolumn{3}{c}{8 layers} & 76.21 &	71.87 \\
\bottomrule

\end{tabular}
\end{table}

\section{Discussion}

In Section~\ref{sec:mainresult}, SleepDIFFormer demonstrated superior performance over existing methods, highlighting its enhanced capability to capture underlying EEG–EOG representations that generalized effectively to unseen datasets. The only exception is the ISRUC dataset, which our method achieved lower ACC with statistical significance. To understand this performance discrepancy more rigorously, we reported the per-class metrics in Table~\ref{tab:ISRUC-per-class-acc}. The analysis reveals that SleepDIFFormer achieved stronger performance on the more challenging stage N1 (the most difficult sleep stage to classify), whereas SleepDG performs slightly better on the other sleep stages, which are relatively easier to classify. Since N1 constitutes only approximately 13.17\% of the ISRUC dataset, while the remaining sleep stages account for 86.83\%, improvements on N1 contribute relatively little to the overall accuracy despite being reflected in the Macro-F1 metric. Balancing performance across challenging N1 stage and other sleep stages remains an important direction for future work. We further validate the contribution of each proposed component in SleepDIFFormer through detailed ablation studies, which collectively demonstrate that they enable our method to achieve consistently better overall performance compared to prior approaches.

    \begin{tableorg}[]
        \centering
        \caption{\added{Per-class evaluation of SleepDIFFormer and SleepDG on the ISRUC dataset in terms of class-wise performance metrics.}}
        \begin{tabular}{lccccc}
        \toprule
            \textbf{Model} & \textbf{Wake} & \textbf{N1} & \textbf{N2} & \textbf{N3} & \textbf{REM} \\
            \hline
            SleepDIFFormer (ACC) & 81.53 & \textbf{36.79} & \textbf{91.22} & 78.55 & 81.10 \\
            SleepDG (ACC) & \textbf{84.53} & 27.21 & 88.21 & \textbf{82.95} & \textbf{86.14} \\
            \hline
            SleepDIFFormer (MF1) & 87.47 & \textbf{43.48} & \textbf{78.17} & 84.33 & \textbf{83.91} \\
            SleepDG (MF1) & \textbf{88.25} & 36.66 & 77.52 & \textbf{85.16} & 83.38 \\
            \bottomrule
            \end{tabular}
        \label{tab:ISRUC-per-class-acc}
    \end{tableorg}

\begin{table}[htbp]
\centering
\caption{\added{Quantitative analysis of differential attention via occlusion experiments. Performance is evaluated when differential attention with the top-$k$, bottom-$k$, near-zero-$k$, random-$k$ differential attention weights are occluded. A baseline with no occlusion ($k=0$) is provided for comparison. The best results are highlighted in bold.}}
\label{tab:top_occlusion_experiment}
\setlength{\tabcolsep}{4.5pt}
\begin{tabular}{lcccccc}
\toprule

 & \multicolumn{2}{c}{\textbf{k = 0 (Baseline)}} & \multicolumn{2}{c}{\textbf{k = 5}} & \multicolumn{2}{c}{\textbf{k = 10}} \\
\cmidrule(lr){2-7}
\textbf{Dataset}& \textbf{ACC} & \textbf{MF1} & \textbf{ACC} & \textbf{MF1} & \textbf{ACC} & \textbf{MF1} \\
\cmidrule(lr){2-7}
\multicolumn{7}{c}{\textbf{Top-k Occlusion}} \\
\hline
\textbf{$\bm{\mathrm{TD}_1}$-SleepEDFx} & \textbf{0.7836} & \textbf{0.7293} & 0.4463 & 0.2808 & 0.3639 & 0.1906 \\
\textbf{$\bm{\mathrm{TD}_2}$-HMC}       & \textbf{0.7500} & \textbf{0.7271} & 0.3700 & 0.3545 & 0.2172 & 0.1479 \\
\textbf{$\bm{\mathrm{TD}_3}$-ISRUC}     & \textbf{0.7808} & \textbf{0.7547} & 0.5205 & 0.4564 & 0.3323 & 0.2334 \\
\textbf{$\bm{\mathrm{TD}_4}$-SHHS-1}    & \textbf{0.7541} & \textbf{0.6759} & 0.4600 & 0.3859 & 0.3500 & 0.2458 \\
\textbf{$\bm{\mathrm{TD}_5}$-P2018}     & \textbf{0.7495} & \textbf{0.7209} & 0.5242 & 0.4654 & 0.3482 & 0.2760 \\
\midrule

\multicolumn{7}{c}{\textbf{Bottom-k Occlusion}} \\
\hline

\textbf{$\bm{\mathrm{TD}_1}$-SleepEDFx} & \textbf{0.7836} & \textbf{0.7293} & 0.7100 & 0.6472 & 0.6858 & 0.6220 \\
\textbf{$\bm{\mathrm{TD}_2}$-HMC}       & \textbf{0.7500} & \textbf{0.7271} & 0.7434 & 0.7086 & 0.7386 & 0.7031 \\
\textbf{$\bm{\mathrm{TD}_3}$-ISRUC}     & \textbf{0.7808} & \textbf{0.7547} & 0.7542 & 0.7148 & 0.7465 & 0.7050 \\
\textbf{$\bm{\mathrm{TD}_4}$-SHHS-1}    & \textbf{0.7541} & \textbf{0.6759} & 0.7380 & 0.6551 & 0.7366 & 0.6539 \\
\textbf{$\bm{\mathrm{TD}_5}$-P2018}     & \textbf{0.7495} & \textbf{0.7209} & 0.6996 & 0.6623 & 0.6785 & 0.6408 \\
\midrule

\multicolumn{7}{c}{\textbf{Near-Zero-k Occlusion}} \\
\hline

\textbf{$\bm{\mathrm{TD}_1}$-SleepEDFx} & 0.7836 & \textbf{0.7293} & \textbf{0.7879} & 0.7277 & 0.7865 & 0.7257 \\
\textbf{$\bm{\mathrm{TD}_2}$-HMC}       & \textbf{0.7500} & \textbf{0.7271} & 0.7360 & 0.7092 & 0.7328 & 0.7059 \\
\textbf{$\bm{\mathrm{TD}_3}$-ISRUC}     & \textbf{0.7808} & \textbf{0.7547} & 0.7745 & 0.7443 & 0.7737 & 0.7436 \\
\textbf{$\bm{\mathrm{TD}_4}$-SHHS-1}    & \textbf{0.7541} & \textbf{0.6759} & 0.7441 & 0.6607 & 0.7444 & 0.6606 \\
\textbf{$\bm{\mathrm{TD}_5}$-P2018}     & \textbf{0.7495} & \textbf{0.7209} & \textbf{0.7495} & 0.7183 & 0.7475 & 0.7164 \\

\midrule
\multicolumn{7}{c}{\textbf{Random-k Occlusion}} \\
\hline

\textbf{$\bm{\mathrm{TD}_1}$-SleepEDFx} & 0.7836 & \textbf{0.7293} & \textbf{0.7889} & 0.7286 & 0.7861 & 0.7249 \\
\textbf{$\bm{\mathrm{TD}_2}$-HMC}       & \textbf{0.7500} & \textbf{0.7271} & 0.7248 & 0.6984 & 0.7110 & 0.6852 \\
\textbf{$\bm{\mathrm{TD}_3}$-ISRUC}     & \textbf{0.7808} & \textbf{0.7547} & 0.7653 & 0.7348 & 0.7558 & 0.7256 \\
\textbf{$\bm{\mathrm{TD}_4}$-SHHS-1}    & \textbf{0.7541} & \textbf{0.6759} & 0.7360 & 0.6510 & 0.7242 & 0.6393 \\
\textbf{$\bm{\mathrm{TD}_5}$-P2018}     & 0.7495 & \textbf{0.7209} & \textbf{0.7496} & 0.7176 & 0.7477 & 0.7152 \\
\bottomrule

\end{tabular}
\end{table}

To further analyze our method, we visualized the differential attention weights on EEG and EOG signals across layers in Fig.~\ref{fig:diffentialattention}. Overall, attention weights tended to be more uniformly distributed in the shallow layers of SleepDIFFormer, while the deeper layers revealed increasingly contrastive and semantically meaningful attention patterns. In the Wake stage, beta waves were observed at several timestamps and were emphasized by the attention weights. From the N1 stage, the EOG showed slow eye movements (SEM), which were emphasized by the attention weights at 1st second. For the N2 stage, sleep spindles appearing around the 6th second were not fully suppressed by SleepDIFFormer, whereas the K-complexes observed at the 7th second were highlighted. In the N3 stage, prominent delta waves, characterized by a larger amplitude and slow oscillations, occurring at the 12th to 13th second were detected and emphasized. Notably, SleepDIFFormer consistently amplified delta-wave activity across nearly all layers, from the shallow to the deep levels. During REM, the EOG segment at the 15th second showed blink features, amplified by SleepDIFFormer in the yellow region of the last layer, and the sawtooth waves (STWs) appeared in the EEG, signifying the characteristics of REM waves. These observations demonstrated the capability of SleepDIFFormer to emphasize relevant characteristics by assigning higher attention weights, while progressively suppressing irrelevant context across layers. In addition to the qualitative analysis, we conducted a quantitative evaluation of MDTA using attention occlusion experiments, as shown in Table~\ref{tab:top_occlusion_experiment}. \added[comment={R2 C6}]{Specifically, during inference, we evaluated performance by masking features associated with the top-$k$ differential-attention weights ($k \in \{0,5,10\}$), thereby removing the contributions of the most highly weighted regions. The results show that \added{occluding} these regions leads to a clear and consistent drop in both ACC and MF1 across all five datasets. This performance degradation indicates that the regions highlighted by MDTA in Fig.~\ref{fig:diffentialattention} correspond to key features on which the model relies for its predictions. Furthermore, the bottom-$k$, near-zero-$k$, and random-$k$ controls produced smaller or minimal effects overall compared with top-$k$ occlusion, although their magnitudes varied across datasets, providing additional support for the qualitative observations. However, the datasets lack expert annotations for specific physiological events, such as sleep spindles, K-complexes, delta waves, and REM eye movements. Thus, the experiment shows that the model relies on highly attended regions that contribute to learning higher-order discriminative representations for sleep-stage classification, while their physiological relevance requires future validation using clinically annotated data.} 
\added[comment={R2 C3}]{We further examine the generalization of our method and SleepDG on unseen subsets of the larger P2018 and SHHS-1 datasets.
As shown in Table~\ref{tab:range}, it achieves generally consistent performance across the different subsets and remains competitive with and outperforms SleepDG in most cases.}

\begin{table}[h]
\centering
\small
\caption{\added{Generalization across different subsets of P2018 and SHHS-1. SleepDIFFormer vs. SleepDG, with the best results highlighted in bold. Rand. ($a$) denotes a randomly selected subset generated using seed $a$.}}
\label{tab:range}
\setlength{\tabcolsep}{3pt}
\setlength{\aboverulesep}{0pt}\setlength{\belowrulesep}{0pt}%
\begin{tabular}{l cc cc |l cc cc}
\toprule
& \multicolumn{4}{c|}{\textbf{P2018}} & & \multicolumn{4}{c}{\textbf{SHHS-1}} \\
\cmidrule(lr){2-5} \cmidrule(lr){7-10}
& \multicolumn{2}{c}{\textbf{SleepDIFFormer}} & \multicolumn{2}{c|}{\textbf{SleepDG}} & & \multicolumn{2}{c}{\textbf{SleepDIFFormer}} & \multicolumn{2}{c}{\textbf{SleepDG}} \\
\cmidrule(lr){2-3} \cmidrule(lr){4-5} \cmidrule(lr){7-8} \cmidrule(lr){9-10}
\textbf{Sub. ID} & \textbf{ACC} & \textbf{MF1} & \textbf{ACC} & \textbf{MF1} & \textbf{Sub. ID} & \textbf{ACC} & \textbf{MF1} & \textbf{ACC} & \textbf{MF1} \\
\midrule
1-150           & \textbf{74.95} & \textbf{72.09} & 74.13 & 69.87 & 1-150     & \textbf{75.41} & \textbf{67.59} & 73.96 & 67.12 \\
201-350         & \textbf{73.55} & \textbf{70.79} & 71.53 & 67.79 & 1001-1150 & \textbf{78.29} & \textbf{70.80} & 75.77 & 69.22 \\
401-550         & \textbf{71.03} & \textbf{68.13} & 70.40 & 65.91 & 2001-2150 & \textbf{74.95} & 67.57 & 74.59 & \textbf{67.79} \\
601-750         & \textbf{72.91} & \textbf{71.17} & 72.25 & 69.09 & 3001-3150 & \textbf{73.70} & \textbf{67.30} & 72.82 & 66.78 \\
801-950         & \textbf{75.11} & \textbf{72.18} & 74.76 & 70.52 & 4001-4150 & \textbf{76.38} & \textbf{70.21} & 75.37 & 69.69 \\
Rand. (0) & \textbf{73.44} & \textbf{71.03} & 72.47 & 68.54 & Rand. (0) & \textbf{76.49} & \textbf{69.79} & 75.96 & 69.78 \\
Rand. (1) & \textbf{73.94} & \textbf{71.32} & 73.49 & 69.38 & Rand. (1) & \textbf{75.73} & \textbf{68.77} & 74.51 & 68.35 \\
\bottomrule
\end{tabular}%
\end{table}

Hence, the results and analyses provided evidence that SleepDIFFormer learned meaningful and interpretable joint EEG \& EOG representations that underlay its superior generalization to unseen datasets, a capability crucial for clinical application. Interpretability was particularly important, as it provided insights into the physiological relevance of the learned representations and allowed us to examine whether the model’s decisions aligned with established sleep science. Such interpretability could increase the clinical trustworthiness of SleepDIFFormer, enabling clinicians to understand not only the classification outcome but also the neural patterns driving the prediction. By enabling robust and automated sleep stage classification, our approach had the potential to reduce reliance on labor-intensive manual scoring of PSG, improve diagnostic efficiency, and facilitate large-scale screening of sleep disorders. This work therefore represented a critical step toward clinically relevant AI systems for sleep quality assessment, diagnosis, and treatment planning.

\section{Conclusion and Future Work} 

In this paper, we propose SleepDIFFormer, a framework for sleep stage classification that learned multi-channel raw EEG and EOG representations while reducing attention noise caused by the non-stationarity of these signals through the use of MDTA. Moreover, we incorporated domain generalization by employing cross-domain feature alignment to minimize domain shifts and extract domain-invariant features for inference on unseen datasets. The proposed model outperformed current state-of-the-art methods on most public sleep staging datasets studied. Furthermore, we provided layer-wise attention visualizations for EEG and EOG inputs, enhancing model interpretability. These findings have implications for advancing automated sleep stage classification and its application to quantifying sleep architecture and detecting abnormalities that impact restorative rest.

While this work focused only on the problem of cross-domain generalization with EEG and EOG signals, future work could investigate the benefits of incorporating additional modalities and leveraging the multi-modal nature of biosignals to better handle the balance between challenging N1 stage with other sleep stages. For example, incorporating EMG could further increase the expressiveness of the model. Another promising direction is to explore feature disentanglement techniques to separate domain-invariant from domain-specific features, ensuring that alignment is applied only to invariant representations. Moreover, extending our method to train on a broader range of EEG datasets could enhance robustness and lay the groundwork for developing an EEG FM capable of generalizing across multiple tasks.




\section*{Declarations}
\subsection*{Funding}
The work was supported by the Ministry of Higher Education Malaysia through the Fundamental Research Grant Scheme (FRGS/1/2023/ICT02/XMU/02/1), Xiamen University Malaysia Research Fund (XMUMRF/2022-C10/IEC E/0039, XMUMRF/2024-C13/IECE/0049), Universiti Malaya Impact Oriented Interdisciplinary Research Grant (IIRG004C-2022SAH), and Geran Penyelidikan Fakulti (GPF017A-2023).

\subsection*{Conflict of interest/Competing interests}
Not applicable.

\subsection*{Ethics approval and consent to participate}
Not applicable.

\subsection*{Consent for publication}
Not applicable.

\subsection*{Data availability}
Public data were analyzed.

\subsection*{Code availability}
Our source code and checkpoint are made publicly available at \url{https://github.com/Ben1001409/SleepDIFFormer}

\subsection*{Author contribution}
Benjamin Wei Hao Chin: Conceptualization, Methodology, Software, Writing - Original Draft. Yuin Torng Yew: Conceptualization, Methodology, Data Curation, Writing - Original Draft. Jia Hong Puah: Conceptualization, Software, Validation, Formal analysis, Writing - Original Draft. Haocheng Wu: Software, Formal analysis, Writing - Review \& Editing, Visualization. Lanxin Liang: Data Curation, Writing - Review \& Editing, Visualization. Chow Khuen Chan: Supervision, Project administration, Funding acquisition. Norita Mohd Zain: Supervision, Project administration, Funding acquisition. Siti Balqis Samdin: Writing - Review \& Editing, Supervision. Sim Kuan Goh: Conceptualization, Resources, Writing - Review \& Editing, Supervision, Project administration, Funding acquisition.

\bibliography{references}

\end{document}